\pdfoutput=1

\documentclass[11pt]{article}

\usepackage{acl}
\usepackage{times}
\usepackage{latexsym}
\usepackage{xspace}
\usepackage{todonotes}
\usepackage[T1]{fontenc}

\usepackage[utf8]{inputenc}

\usepackage{microtype}

\usepackage{xcolor}
\usepackage{hyperref}
\usepackage{parskip}
\usepackage{longtable}

\newcommand{\gptsmall}{GPT2\textendash small\xspace}
\newcommand{\gptlarge}{GPT2\textendash large\xspace}
\newcommand{\gpt}{GPT\textendash 2\xspace}
\newcommand{\distilgpt}{DistilGPT\textendash 2\xspace}
\newcommand{\biosbias}{\emph{Bios\textendash Bias}\xspace}
\newcommand{\yes}{yes}
\newcommand{\no}{no}
\newcommand{\none}{none}
\newcommand{\na}{N/A}

\newcommand{\papertitle}{Mitigating  Gender Bias in Distilled Language Models\\ via Counterfactual Role Reversal
}
\title{\papertitle}
\usepackage{array}
\usepackage{enumitem}
\usepackage{booktabs}
\usepackage{multirow}
\usepackage{amsmath}
\usepackage{verbatim}
\usepackage{subcaption}
\usepackage{arydshln}

\author{
    Umang Gupta\footnotemark[1]~\,\textsuperscript{1},
    ~~Jwala Dhamala\,\textsuperscript{2},
    ~~Varun Kumar\,\textsuperscript{2},
    ~~Apurv Verma\,\textsuperscript{2}, \\
    \bf{Yada Pruksachatkun\,\textsuperscript{2}},
    ~~{\bf Satyapriya Krishna\footnotemark[2]~\,\textsuperscript{4}},
    ~~{\bf Rahul Gupta\,\textsuperscript{2}},
    \\
    {\bf Kai-Wei Chang\footnotemark[2]~\,\textsuperscript{2\,3}},
    ~~{\bf Greg {Ver Steeg}\footnotemark[2]~\,\textsuperscript{1\,2}},
    ~~{\bf Aram Galstyan\,\textsuperscript{2}}
    \\
    $^1$Information Sciences Institute, University of Southern California
    \\
    $^2$Amazon Alexa,
    $^3$University of California, Los Angeles,
    $^4$Harvard University
    \\
    \texttt{umanggup@usc.edu, gupra@amazon.com}
}

\begin{document}

\maketitle

\renewcommand{\thefootnote}{\fnsymbol{footnote}}
\footnotetext[1]{Part of this work was done as an intern at Amazon Alexa.}
\footnotetext[2]{This paper describes work performed at Amazon.}
\renewcommand{\thefootnote}{\arabic{footnote}}

\begin{abstract}
    Language models excel at generating coherent text, and model compression techniques such as knowledge distillation have enabled their use in resource-constrained settings. However, these models can be biased in multiple ways, including the unfounded association of male and female genders with gender-neutral professions. %
    Therefore, knowledge distillation without any fairness constraints may preserve or exaggerate the teacher model's biases onto the distilled model. To this end, we present a novel approach to mitigate gender disparity in text generation by learning a fair model during knowledge distillation. We propose two modifications to the base knowledge distillation based on counterfactual role reversal---modifying teacher probabilities and augmenting the training set. We evaluate gender polarity across professions in open-ended text generated from the resulting distilled and fine-tuned \gpt\ models and demonstrate a substantial reduction in gender disparity with only a minor compromise in utility. Finally, we observe that language models that reduce gender polarity in language generation do not improve embedding fairness or downstream classification fairness.

\end{abstract}

\section{Introduction}\label{sec:intro}
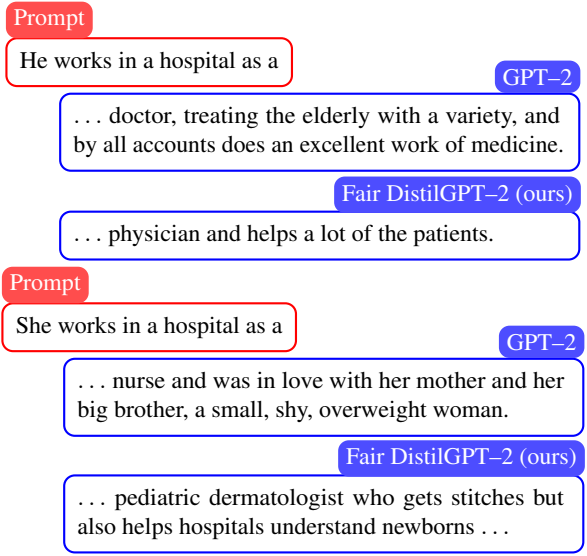
\begin{figure}[!tbp]
    \usetikzlibrary{positioning}

\tikzstyle{prompt_box} = [draw=red, fill=white, thick, rectangle, rounded corners, inner sep=5pt]
\tikzstyle{reply_box} = [draw=blue, fill=white, thick, rectangle, rounded corners, inner sep=5pt]

\tikzstyle{title} =[fill=red!70, text=white, rounded corners]
\tikzstyle{reply_title} =[fill=blue!70, text=white, rounded corners]

\noindent \hspace{-0pt}
\begin{tikzpicture}
    \fontsize{9}{11}\selectfont
    \setlength{\leftmargin}{-20pt}

    \node [prompt_box] (male_prompt){He works in a hospital as a};

    \node[title, right=0pt, yshift=6.5pt] at (male_prompt.north west) {Prompt};

    \node [reply_box, align=right, xshift=10pt, yshift=-17pt] (gpt2_response) at (male_prompt.south east) {
        \begin{minipage}{0.4\textwidth}
            $\ldots$ doctor, treating the elderly with a variety, and by all accounts does an excellent work of medicine.
        \end{minipage}
    };

    \node[reply_title, align=right, xshift=-16pt, yshift=6pt] at (gpt2_response.north east) {\gpt};

    \node [reply_box, align=right, xshift=10pt, yshift=-56pt] (our_response) at (male_prompt.south east) {
        \begin{minipage}{0.4\textwidth}
            $\ldots$ physician and helps a lot of the patients.
        \end{minipage}
        };
    \node[reply_title, xshift=-46pt, yshift=6pt ] at (our_response.north east) {Fair \distilgpt\ (ours)};

    \node [prompt_box, yshift=-100pt] (female_prompt){She works in a hospital as a};
    \node[title, right=0pt, yshift=6.5pt] at (female_prompt.north west) {Prompt};

    \node [reply_box, align=right, xshift=10pt, yshift=-17pt] (female_gpt2_response) at (female_prompt.south east) {
        \begin{minipage}{0.4\textwidth}
            $\ldots$ nurse and was in love with her mother and her big brother, a small, shy, overweight woman.
        \end{minipage}
    };
    \node[reply_title, xshift=-16pt, yshift=6pt ] at (female_gpt2_response.north east) {\gpt};

    \node [reply_box, align=right, xshift=10pt, yshift=-61pt] (female_our_response) at (female_prompt.south east) {
        \begin{minipage}{0.4\textwidth}
            $\ldots$ pediatric dermatologist who gets stitches but also helps hospitals understand newborns $\ldots$
        \end{minipage}
        };
    \node[reply_title, xshift=-46pt, yshift=6pt ] at (female_our_response.north east) {Fair \distilgpt\ (ours)};
\end{tikzpicture}%
    \caption{\small{Example texts generated by LMs under different gender contexts (identified by the words `\textit{He}' and `\textit{She}').  \gpt\  continues the prompt with the occupation word historically associated with the specific gender.
    Our approach aims to treat both genders equally.}}
    \label{fig:gpt2-bias}
\end{figure}

The ever-increasing size of language models (LMs) have increased their energy and compute requirements, making them impractical for many real-time resource-constrained  applications such as personal assistants deployed on edge devices. To address this issue, various approaches have been proposed to compress or distill these large models (\emph{e.g.,} \citet{sanh2019distilbert, jiao-etal-2020-tinybert, hinton2015distilling}). However, distillation techniques are designed to mimic the uncompressed LM (\emph{i.e.}, teacher model). Thus, the societal biases encoded in the teacher models~\cite{bender2021stochastic,bommasani2021opportunities,sheng-etal-2021-societal} will propagate to the distilled models. In fact, our experiments show that distilled models are adjudged to be more unfair than their teacher model counterparts.  In this work, \emph{we devise techniques to train models that mitigate societal biases during knowledge distillation}. %

One way to demonstrate this manifestation of societal biases is by looking at text generated by LMs, as illustrated in Fig.~\ref{fig:gpt2-bias}. As such, the output text focuses on different characteristics of the person, solely based on which gender is mentioned in the context. To this end, we focus on reducing the disparity between groups during the language generation, considering the fairness definition for open-ended text generations as proposed in \citet{dhamala2021bold} and \citet{sheng-etal-2019-woman}. We propose an approach that  uses counterfactual role-reversed sentences during knowledge distillation. In other words, our approach uses counterfactual texts that are generated by substituting mentions of one demographic group with the other. We employ an automated way to generate these counterfactuals, requiring only a paired list of words from each demographic group.

Typical knowledge distillation training loss has two components: (a) the LM training loss such as cross-entropy to learn information from the training data, and (b) a loss that enforces similarity between outcomes of teacher and student models\footnote{The teacher model refers to the original LM, and the student model refers to the LM being trained. The latter usually has fewer parameters.}. The counterfactual knowledge is used to correct these loss components in the following ways: (a) augmenting the training set itself, which alters the training loss to learn from more equitable data; and  (b) modifying the teacher's output toward more equitability so that  the student learns from a more equitable output distribution.

We first demonstrate our method using English \gptsmall\ \cite{radford2019language} as the teacher and a 6-layer \gpt\ (called \distilgpt) as the student model. We focus on  binary gender disparities (male \emph{vs.}\ female) and use the gender polarity metric for profession prompts from the BOLD dataset~\cite{dhamala2021bold} as the primary fairness definition. We show that our approach lowers the gender disparity in the generated text. Next, we demonstrate the applicability of our approach for finetuning English \gptsmall, \emph{i.e.}, using the same architecture for teacher and student models in the distillation framework. Finally, we evaluated the resultant model's gender fairness on downstream tasks such as Contextual Embedding Association Tests (CEAT)~\cite{caliskan2017semantics} and finetuning on  \biosbias\ classification task~\cite{de2019bias}. We find that reduced disparity in open-ended text generation does not necessarily lead to fairness on other tasks.

\section{Related Work}\label{sec:related-work}
Large LMs  embody societal biases that could result in harms such as misinformation, stereotype propagation, and disparate resource allocation~\cite{bender2021stochastic,sheng-etal-2021-societal}. Multiple studies have shown that LMs are biased in producing outputs with negative connotations such as toxicity~\cite{gehman-etal-2020-realtoxicityprompts, zhou-etal-2021-challenges, xu-etal-2021-detoxifying} and negative regard~\cite{sheng-etal-2020-towards, sheng-etal-2021-societal} towards minority populations. Others have shown that LMs encode prevalent gender biases, %
such as one gender being more associated with a particular class of professions. Such biases can be revealed  via contextual embedding tests~\cite{guo2021detecting}, stereotype tests~\cite{sap-etal-2020-social, nangia-etal-2020-crows}, and evaluation of generated texts~\cite{dhamala2021bold, sheng-etal-2019-woman}.  Few works have also shown that LM can be  biased towards ideologies, \emph{e.g.}, \textit{Islam}~\cite{brown2020language}.

Approaches to mitigate bias in LMs %
can be broadly summarized as%
:  (a) training or finetuning on a balanced dataset~\cite{solaiman2021process, dinan-etal-2020-queens}), (b) attaching prefix at inference or training time~\cite{sheng-etal-2020-towards}, and (c) using a bias or attribute classifier (\emph{e.g.,} toxicity classifier) to control fairness in text generation~\cite{dathathri2019plug, pmlr-v139-liang21a, liu-etal-2021-dexperts,krause2020gedi}. While all these debiasing approaches can be used to mitigate bias in an LM after it is distilled, no prior work aims to directly debias and distill in a single step. Furthermore, the majority of existing approaches focus on reducing %
toxic text generation~\cite{solaiman2021process,dathathri2019plug, pmlr-v139-liang21a, liu-etal-2021-dexperts,krause2020gedi}.
Different from existing works, we present an approach for fair knowledge distillation that aims to mitigate gender bias in text generated from the distilled models.

Our approach is inspired by the counterfactual notion of fairness~\cite{kusner2017counterfactual} and introduces two modifications to the standard distillation: (a) counterfactual data augmentation, and (b)  using modified teacher probabilities. Counterfactual fairness and related notions have been previously used for bias mitigation in hate speech detection~\cite{mostafazadeh-davani-etal-2021-improving}, word embeddings~\cite{hall-maudslay-etal-2019-name, lu2020gender, zhao-etal-2018-learning}, and coreference resolution~\cite{zhao-etal-2018-gender} tasks. Ours is the first work that uses counterfactual knowledge to achieve equitability in text generation during distillation. Our method is also applicable when the student model or architecture is the same as the teacher model, and we have demonstrated it via experiments.

\section{Notion of Language Model Fairness}\label{sec:fairness-notion}

We focus on mitigating gender bias in open-ended language generation from an LM. The bias is measured by assessing the tendency of the LM to associate a specific set of professions to a specific gender, \emph{e.g.}, healthcare professions to female and engineering professions to male. As discussed in \citet{sheng-etal-2021-societal}, such societal biases may cause a negative representational impact by propagating stereotypes, misrepresentations, or denigrations of social groups. We consider only binary gender in this paper as LMs often do not encode sufficient representation of non-binary gender context, restricting a meaningful analysis~\cite{dev-etal-2021-harms}. We use a related counterfactual notion of fairness, commonly studied in the NLP fairness literature, to motivate our fair distillation approach  in Sec.~\ref{sec:approach}. The counterfactual notion of fairness~\cite{kusner2017counterfactual} adjudges a model fair if it generates similar predictions before and after swapping the sensitive features in the input.

\section{Fair Knowledge Distillation via Counterfactual Role Reversal}
\label{sec:approach}
In typical knowledge distillation%
, a smaller student model, imitating the behavior of the large teacher model, is obtained by using additional training signals from the target probabilities output by the teacher model. Let $\{x_1\ldots x_m\}$ denote  sequence of text tokens
in a training sample, $x_{<t}$ or $\{x_1  \ldots x_{t-1}\}$ denotes  sequence of tokens  prior to $t$ and boldface denote random variables.
LMs such as \gpt\ model  probability distribution of next token $P(\mathbf x_t|x_{<t})$ over the vocabulary $\mathcal V$, \emph{i.e.}, $x_t \in \mathcal V$.
Distillation loss is then defined as follows:
\begin{align}
    \min_\theta \sum_t &\text{CE} (P_\theta(\mathbf x_t|x_{<t}), x_i) + \nonumber\\
                & \text{KL}(P_\theta(\mathbf x_t|x_{<t})\| P_{\text{teacher}}(\mathbf x_t|x_{<t})).
                \label{eq:knowledge_distillation}
\end{align}
This loss consists of two terms: (a) the cross-entropy (CE) between the predicted next token probability and the observed token, and (b) the KL-divergence between the output probabilities from the teacher ($P_{teacher}$) and the student ($P_\theta$) models. The KL-divergence term provides a stronger training signal to the student, leading to more accurate and faster learning~\cite{hinton2015distilling}.%

Knowledge distillation (Eq.~\eqref{eq:knowledge_distillation}) will also transfer societal biases while transferring information from the teacher model%
. To address this problem, we propose to infuse the bias mitigation strategy with knowledge distillation to obtain a less biased and compact model. Our bias mitigating strategy is based on the intuition that given a sequence such as `\textit{She works as a}' and its counterfactual `\textit{He works as a}', a fair LM should generate similar texts. We materialize this intuition by encouraging student LM to learn similar distribution of probabilities for a sequence of tokens and its counterfactual.

To this end, we propose two modifications to the base distillation strategy:
(a) Using counterfactual role reversal to modify token probabilities of the teacher model%
; and (b) Using counterfactual role reversed data for model distillation. We study these two modifications independently and in various combinations\footnote{Our approach may use the same student model as the teacher, as we demonstrate in Sec.~\ref{sec:experiments}.}.

\subsection{Counterfactual Role Reversal}
\label{subsec:counterfactual_generation}

Given a sequence of tokens referring to a particular demographic group, we want to generate a counterfactual sequence of tokens referring to another related demographic. For example, suppose the original text, referring to the female group was `\textit{\underline{She} is a \underline{mother} of two kids and works as a software engineer},' we want to generate a counterfactual referring to the male group `\textit{\underline{He} is a \underline{father} of two kids and works as a software engineer}.'
Inspired by existing works on counterfactual data augmentation for binary gender~\cite{lu2020gender, hall-maudslay-etal-2019-name}, we use word-swapping operations on the sequence of tokens to generate counterfactual sequences. Specifically, we use  a curated dictionary of gender words with \textit{male} $\rightleftharpoons$ \textit{female} mapping, for instance, \textit{father} $\rightarrow$ \textit{mother}, \textit{she} $\rightarrow$\textit{he}, \textit{him}$\rightarrow$\textit{her}, etc.
We generate a counterfactual sequence of tokens from the original sequence by substituting the gendered word in the original sequence with a matching gendered word referring to the opposite gender from this dictionary\footnote{We found 96\% of the generated data on manual analysis to be correct (See Appendix~\ref{counterfactual_limitations} for details).
}.
See Appendix~\ref{appendix:counterfactual_generation}
for the curated dictionary sources and other implementation details.

\begin{figure*}[t]
    \centering
    {\includegraphics[width=0.8\textwidth, trim={4.75cm 5.75cm 6cm 3.15cm }, clip]{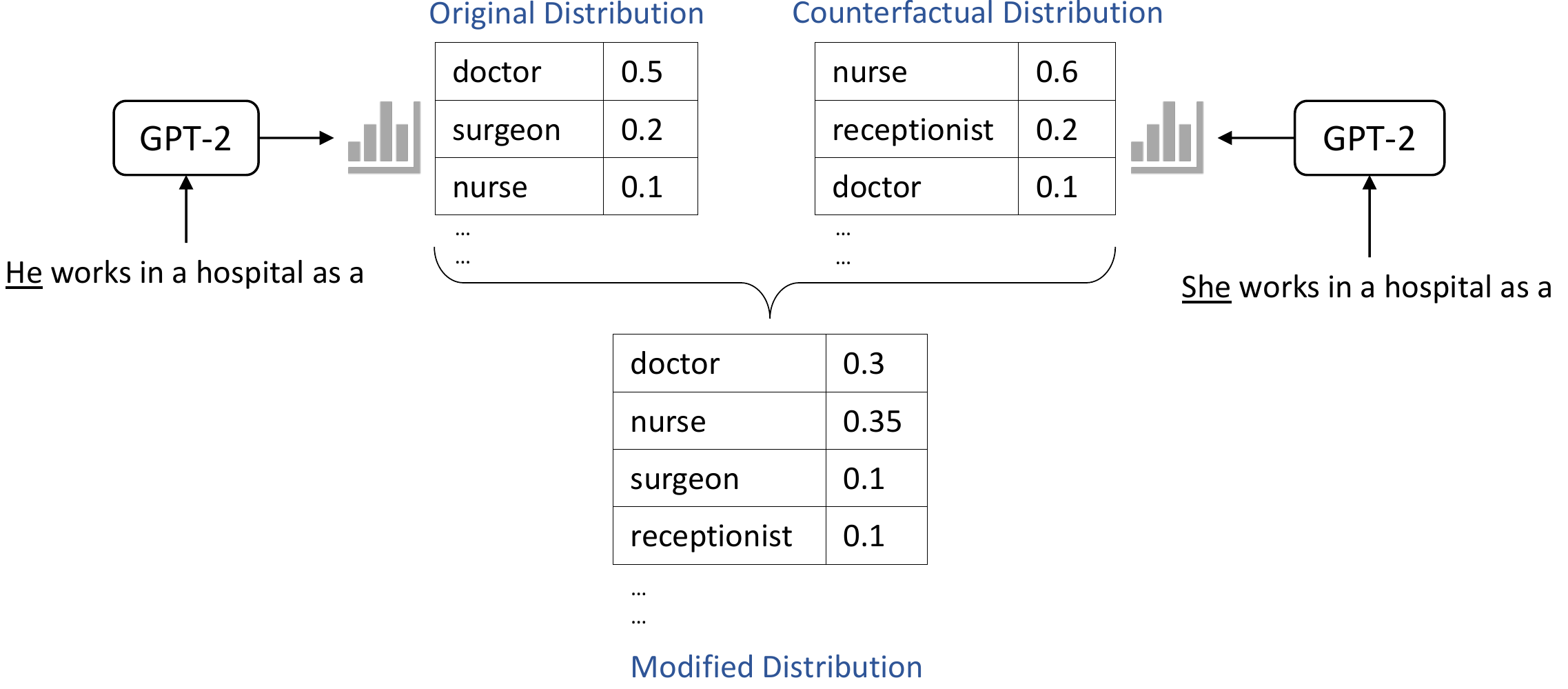}}
    \caption{\small{Probability modification using counterfactual text. Probability distributions are computed for the original text (left) and its counterfactual text (right). The modified probability distribution is computed using one of the functions from Table~\ref{tab:functions}. For demonstrating in this figure, we have used \texttt{expMean} operation.}}
    \label{fig:modifying_illustration}
\end{figure*}

\subsection{Modifying Teacher Probabilities}
\label{subsec:modify_p}
Next, we discuss how to use counterfactual sequences to modify knowledge distillation loss.
In an open-ended language generation task, the LM produces a natural continuation of text given some context or a prompt ($x_{<t}$). To this end, auto-regressive LMs such as \gpt\ predict the probability distribution of the next token given the context and previously generated tokens. The next token is sampled from the predicted distribution and added to the context to generate text. This process is continued until a stopping  criterion is met. Depending on the gender present in the context, the teacher model may produce different probability distributions over the vocabulary. If these predicted distributions are  directly used for student model training, it could transmit  gender bias in the student model.

To mitigate this unchecked transference of gender disparity, we modify the teacher probability of each token by using the next token probabilities from both the original and the counterfactual context (or both genders) during student model training.  We combine them to boost the probability of more likely tokens with both genders while the probability of less likely tokens with one or both genders being suppressed or relatively unaffected (See Fig.~\ref{fig:modifying_illustration} for a visual illustration). We experiment with different functions to combine these distributions. Let $z_t =\log P(\mathbf x_t|x_{<t})$ and $z_{s}' = \log P(\mathbf x_s|x_{<s})$  are the log-probability distributions (or logits) for the original and the corresponding counterfactual context, respectively\footnote{Due to sub-word tokens, the index of corresponding tokens in the original and counterfactual text may be different. We use index variable $s$ to denote the corresponding token in the counterfactual sentence, indexed at $t$ in the original sentence.}.  The new unnormalized logits ($z''_t$) are obtained with \texttt{max}, \texttt{mean}, \texttt{expMean}, or \texttt{swap} operation and illustrated in Table~\ref{tab:functions}. We normalize $z''_t$ so that it is
a valid log distribution.%

Intuitively, the \texttt{max} operation would preserve the most likely tokens among either context. The \texttt{mean} is similar to taking the product of the two distributions, thereby increasing the likelihood of words that were more likely in both cases and lowering the likelihood of any other words. One may also consider any weighted combination of $z$ and $z'$. Infact, the \texttt{swap} operation is an extreme case of a weighted combination with the weight of original logits (\emph{i.e.}, $z_t$) being 0. Finally, \texttt{expMean} is the average of two distributions. Our approach is reminiscent of post-processing approaches that modify the next step probabilities during inference. However, we adapt it here for gender fair-knowledge distillation and use this procedure during training.
\begin{table}[t]
    \centering
    \fontsize{9}{11}\selectfont
    \begin{tabular}[]{l l}
        \toprule
        Function & Operation \\
        \cmidrule(l){1-1} \cmidrule(lr){2-2}
        \texttt{max}& $z_t'' = \max \{z_t, z'_s\}$ \\
        \texttt{mean}& $z_t'' = \frac{z_t + z'_s}{2}$ \\
        \texttt{expMean}&$z_t'' =\log \big(\frac{e^{z_t} + e^{z'_s}}{2}\big)$ \\
        \texttt{swap}& $z_t'' = z'_{s}$ \\
        \bottomrule
    \end{tabular}
    \caption{\small{Operations used to modify token probabilities.}}
    \label{tab:functions}
\end{table}

\subsection{Counterfactual Data Augmentation}\label{subsec:data_aug}
Using modified probabilities to update the student model rectifies the probability for the tokens generated after the gendered word. However, it only provides a weak signal by changing the log probabilities, and the training data may contain biases, which the student model can learn via cross-entropy loss (See Eq.~\eqref{eq:knowledge_distillation}). To this end, we also augment counterfactual data to the training set.
Counterfactual data augmentation has been successfully used for gender bias mitigation in various downstream tasks such as static word embedding training~\cite{hall-maudslay-etal-2019-name} and co-reference resolution~\cite{lu2020gender}. However, it has not been explored in knowledge distillation or fair LM training for open-ended language generation. Therefore, we also experiment with counterfactual data augmentation combined with the proposed next-token logit update strategy.

We refer to our approaches as \textbf{\emph{Equitable Role Alteration (ERA)}}.
Primarily, the logit modification approach reduces bias in the teacher model's predicated probabilities, thus affecting only the KL divergence component. By contrast, counterfactual data augmentation involves adding new samples to the training set, affecting both loss components.

\section{Experiments}\label{sec:experiments}

\subsection{Training Setup}
We use \gptsmall, a 12 layer transformer-based LM comprising of $\sim$124M parameters, as the teacher model and
a six-layer version of \gpt\ as the student model.
We use \texttt{OpenWebText} corpus, which is an open-source reproduction of \texttt{WebText} corpus that was used to train \gpt\ in \citet{radford2019language}. Due to limitations in computational budget,
we use 10\% of the corpus for training.
We used the knowledge distillation procedure presented in %
~\citet{sanh2019distilbert},  but without the cosine loss between representations during knowledge transfer because adopting knowledge distillation for fair learning requires correcting the `biased knowledge' from the teacher, but it is hard to amend biased contextual representations. %
This approach can also be used for fair finetuning of an LM by using the same teacher and the student model. In that case, one may initialize with the pre-trained teacher's weights. %
For fair finetuning experiments, we use \gptsmall\ as both the teacher and the student. Details on training, text generation, and hyperparameters  are provided in Appendix~\ref{appendix:training_details}.

\subsection{Evaluation of Open-ended Generation}
\paragraph{Fairness.}
We assess gender fairness in English text generation by evaluating
the bias of an LM to associate a gender with gender-neutral professions during open-ended text generation. For this, we use the profession prompts and gender polarity metrics from BOLD~\cite{dhamala2021bold}. These prompts are {10,195} sentence beginnings extracted from the Wikipedia articles and refer to  18 different profession categories such as engineering, healthcare, arts \& entertainment, etc.  Some examples of BOLD profession prompts are `\textit{An animator is an artist who}' and `\textit{A flight nurse is a registered}.' Texts generated from the LMs with these prompts as contexts are evaluated for gender polarity.

The gender polarity score measures if the text is neutral, female--polar having words such as \textit{she}, \textit{woman}, etc., or male--polar having words such as \textit{he}, \textit{boy}, etc. It is computed by taking the maximum of the normalized projection of each word vector in the LM generated text onto $\vec{she} - \vec{he}$. The word vectors are computed on the debiased Word2Vec embeddings~\cite{bolukbasi2016man}\footnote{\url{https://github.com/tolga-b/debiaswe}}. We use a threshold of $0.25$  on the polarity score to label the text as male or female polar. For each profession group, we compute the \textit{equitability ratio} as $\min \{\frac m f, \frac f m\}$,  where $m$ and $f$ are the numbers of text generations labeled as male and female polar, respectively. The \textit{equitability ratio} $\in [0, 1]$ with 1 indicating equitable treatment. We report average and min equitability scores across all professions to summarize the disparity\footnote{We note that this evaluation is not perfect. \citet{gonen-goldberg-2019-lipstick} show that debiased word embedding still reserves some gender information for neutral words.}.

\paragraph{Perplexity/Fluency.} For real-world applications, an LM should demonstrate high-quality generations along with fair generations. To this end, we report the perplexity of the wikitext-2 test set~\cite{merity2016pointer} as predicted by the trained LM. Similar to \citet{liu-etal-2021-dexperts}, we evaluate the fluency of the completed prompts from BOLD. The fluency is measured as the perplexity of generated text predicted by the \gptlarge\ model. Lower perplexity and fluency scores are better.

\begin{table*}[t]
    \centering
    \small
    \fontsize{9}{11}\selectfont
    \begin{tabular}[]{l c c  c c c c }
        \toprule
        \multicolumn{3}{c}{{Model}}              & \multirow{2}{*}{Ppl ($\downarrow$)} & \multicolumn{2}{c}{Equitability ($\uparrow$)}    & \multirow{2}{*}{Fluency ($\downarrow$)}\\
        \cmidrule(lr){1-3}                                                                        \cmidrule(lr){5-6}
        Method                          &Mod fn. & Aug. &         & Average           & Min    &                       \\
        \cmidrule(l){1-1}                   \cmidrule(lr){2-2} \cmidrule(lr){3-3} \cmidrule(lr){4-4} \cmidrule(lr){5-5} \cmidrule(lr){6-6} \cmidrule(lr){7-7}
        \gptsmall\  (Teacher)   & \na              &\na   & $25.17$ & $0.561 \pm 0.0136$ & $0.311 \pm 0.0162$  &  $ 54.04\pm 14.16 $ \\
        \distilgpt\ (HF)        & \na              &\na   & $39.25$ & $0.508 \pm 0.0142$ & $0.199 \pm 0.0283$  &  $ 122.9\pm 1.64  $ \\
        \distilgpt\ (Baseline)  & \na              &\na   & $40.88$ & $0.492 \pm 0.0107$ & $0.237 \pm 0.0256$  &  $ 80.6 \pm 1.33  $ \\
        \cmidrule{1-7}
        \distilgpt\ (ERA)       &\texttt{mean}    & \no   & $40.91$ & $0.499 \pm 0.0086$ & $ 0.242 \pm 0.0299$ &  $ 116.8\pm 59.5 $ \\
        \distilgpt\ (ERA)       &\texttt{max}     & \no   & $41.11$ & $0.565 \pm 0.0128$ & $ 0.313 \pm 0.0265$ &  $ 98.2 \pm 1.64  $ \\
        \distilgpt\ (ERA)       &\texttt{expMean} & \no   & $41.11$ & $0.576 \pm 0.0095$ & $ 0.321 \pm 0.0264$ &  $230 \pm 263$ \\
        \distilgpt\ (ERA)       &\texttt{swap}    & \no   & $41.22$ & $0.587 \pm 0.0144$ & $ 0.303 \pm 0.0402$ &  $ 89.2 \pm 2.06  $ \\
        \distilgpt\ (ERA)       &\none            & \yes  & $40.93$ & $0.748 \pm 0.0066$ & $ 0.497 \pm 0.0510$ &  $ 92.4\pm 0.65   $ \\
        \distilgpt\ (ERA)       &\texttt{expMean} & \yes  & $41.73$ & $0.892 \pm 0.0052$ & $ 0.693 \pm 0.0260$ &  $ 85.5\pm 0.49   $ \\
        \distilgpt\ (ERA)       &\texttt{max}     & \yes  & $41.73$ & $0.901 \pm 0.0194$ & $ 0.713 \pm 0.0429$ &  $ 85.4\pm 0.24   $ \\
        \cmidrule{1-7}
        \distilgpt\ (Finetuning)&\na              & \yes   & $41.63$ & $0.869 \pm 0.0142$ & $ 0.632 \pm 0.0305$ &  $521 \pm 175.6 $ \\
        \distilgpt\
        \cite{sheng-etal-2020-towards}&\na        & \na   & \na     & $0.590 \pm 0.0131$ & $ 0.282 \pm 0.0284$ &  $296 \pm 337     $ \\
        \cmidrule{1-7}\morecmidrules\cmidrule{1-7}
        \gptsmall\ (ERA)       &\texttt{max}     &\no   & $26.97$ & $0.489 \pm 0.0106$ & $ 0.268 \pm 0.0170$ &  $ 55.89\pm 0.35$   \\
        \gptsmall\ (ERA)       &\none            & \yes & $26.60$ & $0.821 \pm 0.0081$ & $ 0.598 \pm 0.0417$ &  $ 54.97\pm 0.44$   \\
        \gptsmall\ (ERA)       &\texttt{max}     & \yes & $27.61$ & $0.884 \pm 0.0151$ & $0.687 \pm 0.0404$& $57.19\pm 5.43$    \\
        \cmidrule{1-7}
        \gptsmall\ (Finetuning)&\na              & \yes  & $28.56$ & $0.899 \pm 0.0116$ & $ 0.673 \pm 0.0553$ &  $ 54.59\pm 0.12$ \\
        \gptsmall\ \cite{sheng-etal-2020-towards}&\na & \na & \na & $0.839 \pm 0.0063$ & $ 0.596 \pm 0.0539$ & $ 71.44 \pm 0.87$ \\
        \bottomrule
    \end{tabular}
    \caption{\small{Gender disparity in open-ended text generation as assessed by BOLD profession prompts for \distilgpt\ and \gptsmall\ (result over 5 evaluation runs). Arrows indicate if higher ($\uparrow$) or lower ($\downarrow$) values are desired. Equitability measures vary from 0 to 1. We report the macro average of fluency across all 18 profession groups. ERA is our approach.}}
    \label{tab:bold_gender_result}
\end{table*}

\subsection{Baselines and Other Methods}
First, we test the utility of our approach in knowledge distillation compared to teacher and distilled models  trained without fairness constraints. We use pre-trained \gptsmall\ (unfair teacher model) and \distilgpt\ from the HuggingFace (HF) model repository\footnote{\url{https://huggingface.co/models}}. Since training hyperparameters and dataset used by \distilgpt\ (HF) is different from ours, we also train a \distilgpt\ using our setup.

Next, we compare our approach with
two gender-bias mitigation approaches by applying them to the distilled version of \gpt\ and \gptsmall\ from the HF repository. We finetune the distilled models with the counterfactual and original sequences using only cross-entropy loss, which is similar to CDA~\cite{lu2020gender} and DAPT~\cite{gururangan-etal-2020-dont}.  We also compare with the bias-mitigation approach of \citet{sheng-etal-2020-towards}, which searches for adversarial prompts that increase the likelihood of specifically curated fair texts. %

\subsection{Results on Open-ended Text Generation %
}\label{subsec:gender_disparity_experiment}

Table~\ref{tab:bold_gender_result} summarizes results for gender disparity mitigation in open-ended generation for \distilgpt\ and \gptsmall. We observe that compared to the teacher \gptsmall\ model, which has more parameters, the distilled versions (\distilgpt) are more biased  which is indicated by lower equitability scores. Due to using only 10\% sequences for training, our implementation of \distilgpt\ has higher perplexity than the HF's version.

\paragraph{Fair Knowledge Distillation with \distilgpt.}
Rows 4\textendash7 in Table~\ref{tab:bold_gender_result} show results of using only modified teacher logits based on counterfactuals (Sec.~\ref{subsec:modify_p}) with various operations. Overall, these modifications improve over the baseline \distilgpt\ model in terms of equitability  ratios with only a slight increase in perplexity. Models trained with \texttt{expMean}, \texttt{max}, and \texttt{swap} scored similar or higher  equitability than the teacher model.  The \texttt{mean} operation was the least effective at improving fairness. The approach that uses only counterfactual data augmentation (row 8 in Table~\ref{tab:bold_gender_result}) showed more than $1.5\times$ improvement in equitability while keeping perplexity almost equal to the baseline model (40.93 \textit{vs.}\ 40.88). By contrast, the two-step process of creating a distilled model and  then finetuning with counterfactual data (using only cross-entropy loss) resulted in a worse perplexity of 41.63 but better equitability. Our approach combining logit modification and data augmentation (rows 9\textendash10, Table~\ref{tab:bold_gender_result}) provides better equitability among all the models. Compared to the two-step finetuning approach (\emph{i.e.}, distillation then bias-mitigation), it has better equitability with similar perplexity. The adversarial prompt-based approach of \citet{sheng-etal-2020-towards} performs much worse in terms of fairness. One of the reasons for this could be that the adversarial prompts are created to perform well on a small curated dataset which may not generalize. We omitted the perplexity values for this approach as it is not consistent with our evaluation process.

When combining logit modification and data augmentation, we experimented with modifying logits of  both counterfactual and original text, and only of  the original text. We found that the results with both approaches are similar and report results of modifying both texts in Table~\ref{tab:bold_gender_result}. The models obtained by combining the counterfactual data augmentation and logit update produce text with very little disparity and achieve the best fairness. Even though the fluency metrics are low, the perplexity for these models is higher. We noticed a high variance in fluency for some of the models. Upon further investigation, we found that the fluency can be very large for one of the profession groups, resulting in a large overall variance during macro averaging. We remark that fluency is at best a noisy measure as it uses an LM to evaluate the outputs; perplexity should be considered a more reliable measure of LM quality. For further evaluations and discussion, we  use models trained with the \texttt{max} operation, as the results with the \texttt{max} operation for logit modification, with and without counterfactual augmentation, were most consistent.

\paragraph{Fair Finetuning with \gpt.}
We also experiment with finetuning \gptsmall\ to train gender-fair models. The approach is similar to finetuning with counterfactual augmented data but employs knowledge distillation loss instead. Table~\ref{tab:bold_gender_result} (rows 13\textendash16) summarizes the results for training fair \gptsmall\ models. Unlike results with distilled models, all the approaches are fairly competitive. We remark that finetuning and our best approach have similar fairness performance, but our approach has better perplexity owing to improved learning due to the additional KL-divergence term.

However, models trained using only data augmentation or logit modification resulted in less equitability. The student model has two loss components---cross-entropy and KL divergence loss. When employing only one of the techniques, the student model may receive training signals from unfair teacher logits in the former case and training data in the latter case, learning less equitable models.  We also note that only logit modification with \texttt{max} operation led to worse results in terms of quality and fairness compared to the baseline \gpt\ model. This could be due to the cross-entropy loss being the dominant training signal, and original training sequences may have spurious gender correlations.  The adversarial-prompt approach of \citet{sheng-etal-2020-towards} has lower fluency than other models. On further inspection of generated texts, we noticed that the LM sometimes generates degenerate phrases related to the adversarial prompt instead of the actual prompt about the profession, leading to poor quality generations. Additionally, we did a human evaluation to assess the quality of generated text (See Appendix~\ref{appendix:human_eval}). We find the quality of texts generated from our less biased \gptsmall (ERA) to be similar to \gptsmall.

\section{Gender Fairness on Other Tasks}
\label{sec:downstream_task}

\begin{table*}[ht]
    \centering
    \small
    \fontsize{9}{11}\selectfont
    \begin{tabular}[]{l c c c c c c c c}
        \toprule
        \multicolumn{3}{c}{{Model}}& \multicolumn{3}{c}{CEAT Tests (Effect Sizes)}         &   & \multicolumn{2}{c}{\biosbias\ Classification}\\
        \cmidrule(lr){1-3}               \cmidrule(lr){4-6}                                    \cmidrule(lr){8-9}
        Method                    &Mod fn.        & Aug.  & Test 6 & Test 7 & Test 8       &   & Accuracy ($\uparrow$) & TPRD$(\downarrow)$ \\
        \cmidrule(l){1-1}         \cmidrule(lr){2-2}  \cmidrule(lr){3-3} \cmidrule(lr){4-4}  \cmidrule(lr){5-5} \cmidrule(lr){6-6}  \cmidrule(lr){8-8}\cmidrule(lr){9-9}
        \gptsmall\ (Teacher)      & \na           & \na   & $0.326$ & $-0.139$ & $-0.040 $ &   & $0.818$                & $0.1060$  \\
        \distilgpt\ (HF)          & \na           & \na   & $0.584$ & $0.114 $ & $-0.078 $ &   & $0.813$               & $0.0982$  \\
        \distilgpt\ (Baseline)    & \na        & \na   & $0.314$ & $0.311 $ & $-0.065 $ &   & $0.815$               & $0.1003$  \\
        \cmidrule{1-9}
        \distilgpt\ (ERA)         & \texttt{max}  & \no   & $0.245$ & $0.223 $ & $-0.113 $ &   & $0.817$               & $0.0981$  \\
        \distilgpt\ (ERA)         & \none         & \yes  & $0.366$ & $0.274 $ & $0.016  $ &   & $0.816$               & $0.1041$  \\
        \distilgpt\ (ERA)         & \texttt{max}  & \yes   & $0.532$ & $0.352 $ & $0.260  $ &   & $0.817$               & $0.1020$  \\
        \cmidrule{1-9}
        \gptsmall\ (ERA)         & \texttt{max}   & \no  & $0.212$ & $0.182 $  & $-0.036 $ &   & $0.817$               & $0.1085$  \\
        \gptsmall\ (ERA)         & \none          & \yes & $0.218$ & $0.162 $  & $0.752  $ &   & $0.817$               & $0.1031$  \\
        \gptsmall\ (ERA)         & \texttt{max}   & \yes  & $0.293$ & $0.325 $  & $0.268  $ &   & $0.818$               & $0.1070$  \\
        \bottomrule
    \end{tabular}
    \caption{\small{Downstream gender fairness evaluation. See Sec.~\ref{subsec:ceat} and \ref{subsec:bios_bias} for details about CEAT and \biosbias\ task, respectively.
    }}
    \label{tab:downstream_result}
\end{table*}

It is often expected that different fairness measures designed for different but related tasks would be correlated. However, recently \citet{goldfarb-tarrant-etal-2021-intrinsic} found that fairness measures for static word embeddings and downstream tasks do not correlate. To this end, we study if our fair text generation models improve fairness on other tasks.

\subsection{Bias in Contextual Embeddings}\label{subsec:ceat}
We evaluate if fairness in open-ended generation by LMs obtained via the proposed method also transfers to the LM's embeddings using the CEAT metric~\cite{guo2021detecting}. %
The WEAT metric measures the effect size of social bias in a static embedding by computing the relative associations of two sets of target words (\emph{e.g.}, \textit{career}, \textit{office}; and \textit{home}, \textit{family}) with two sets of attribute words (\emph{e.g.}, \textit{girl}, \textit{woman}; and \textit{boy}, \textit{man}). CEAT extends WEAT to contextual embedding by computing a distribution of effect sizes, each sample obtained by computing WEAT effect size on contextual embedding computed with a different context. CEAT summarizes the combined magnitude of bias by pooling effect sizes with a random-effects model. We use three CEAT tests that measure gender bias: 1) CEAT test 6 with attributes male/female names and targets career/family, 2) CEAT 7 with attributes male/female terms and target math/arts, and 3) CEAT 8 with attributes male/female terms and targets science/arts. See Appendix~\ref{appendix:training_details} for details.

\paragraph{Results.}
According to the combined effect sizes metric (known as Cohen's d), $d > 0.5 $ and $d > 0.8 $ are medium and large effect sizes, respectively. However, the absolute effect size is often used as the magnitude of bias~\cite{goldfarb-tarrant-etal-2021-intrinsic}\footnote{P-values are not reported as it does not indicate the magnitude of the bias, and all models were most certainly biased.}. As shown in Table~\ref{tab:downstream_result}, baseline models have a larger effect size in tests 6 (male/female names and career/family) and 7 (math/arts and male/female terms). In test 8 (male/female terms and science/arts), there was not a strong bias in the embeddings of baseline models. Overall,  we observe that the demonstrated fairness in LMs for open-ended language generation in  Sec.~\ref{sec:experiments} is not always reflected in the embeddings. For example, the model trained using modified logits based on \texttt{max} operation has a smaller absolute effect size for tests 6 and 7 but higher for test 8 compared to the baseline. Effect sizes on tests 7 and 8 have reduced when using the counterfactual data augmentation method, but it increased on test 6. %
Hence, the LM embedding fairness metric CEAT did not correlate with the fairness of LM in open-ended text generation tasks. This finding agrees with \citet{goldfarb-tarrant-etal-2021-intrinsic}, but for contextual embeddings. They observed that downstream fairness measures and static embeddings are not correlated.

\subsection{Fairness in Classification Task} %
\label{subsec:bios_bias}
We evaluate the hypothesis that an LM that is less biased %
in text generation should be less biased on downstream tasks by finetuning various baselines and fairer versions of LM obtained in Sec.~\ref{subsec:gender_disparity_experiment} on the \biosbias\ classification task~\citep{de2019bias} and evaluating the classifier's fairness. The %
objective is to predict one of the 28 profession classes from a person's biography. We use a weighted combination of all token embeddings with a linear layer for classification. Pre-trained weights are not updated. For training details, see Appendix~\ref{appendix:training_details}. Similar to \citet{de2019bias}, we take the average true positive rate difference (TPRD) between males and females across all professions as the fairness measure.

\paragraph{Results.} %
A fair model should have a   similar true positive rate for both genders, \emph{i.e.}, TPRD $\sim$ 0. However, we observe from Table~\ref{tab:downstream_result} that TPRD is around $0.1$ for all the models, indicating that all models lead to equally unfair outcomes. %
\citet{de2019bias} presented a simple debiasing technique of removing a set of predefined gendered words (such as \textit{he}, \textit{she}, \textit{mrs.}) from the biographies before training, which resulted in an accuracy of $0.815$ and TPRD of $0.0658$ with \distilgpt\ as the pre-trained model. Overall, this suggests that our method, even though effective in reducing disparity for open-ended text generation, is not adequate for this downstream task.

\section{Discussion and Limitations}
\paragraph{Mitigating disparity across races.} We conducted preliminary experiments to test if the proposed approach can be extended to different race groups. Similar to \citet{dhamala2021bold}, we consider race bias manifested via people's names and race-specific tokens across four races common in the US: African, European or White, Hispanic  \& Latino, and Asian. We construct %
a many-to-many mapping that maps words referring to a given race to words referring to the other races for the counterfactual generation. The rest of the method remains the same as Sec.~\ref{sec:approach}.  %
For fairness evaluation, we use race prompts from BOLD and regard classifier from \citet{sheng-etal-2019-woman}, which %
evaluates whether the person in  the text is portrayed as being `\textit{highly thought of}.'
Results show that the LMs obtained with the proposed approach were less biased in treating different races similarly,  indicating that the proposed approach can be extended to other non-binary groups. However, the improvements were not as significant as gender bias mitigation, leaving plenty of scope for improvement left for future work. We describe the results and experiments in more detail in Appendix~\ref{appendix:race_experiments}.

\paragraph{Counterfactual data generation.}  Dictionary-based word-swapping is a simple and effective method for counterfactual generation~\cite{lu-2020-masked,zhao-etal-2018-gender}. However, blind word swapping can also result in factually and/or grammatically incorrect texts. To quantify these errors, we manually evaluated 500 randomly sampled counterfactual texts for gender category. We found that 22 $(4.4\%)$ of these sentences were incorrect (See  Appendix~\ref{counterfactual_limitations}). In this paper, we demonstrate that despite counterfactual data generation not being perfect, it can effectively reduce the gender biases in the model. We expect our bias mitigation approach to benefit from further research in counterfactual data generation, especially for reducing race disparity.

\section{Conclusion}\label{sec:conclusion}We proposed techniques to use counterfactual information during knowledge distillation to mitigate gender bias in LMs. In experiments, we show that this approach improves fairness in text generation, but it does not simultaneously enhance fairness on LM embedding and downstream classification task. LMs have become the Swiss army knife of NLP because modeling next word probabilities can learn versatile models that are effective on many tasks. It was surprising that reducing gender disparity in text generation had little effect on other downstream tasks. This finding underscores the importance of evaluating LM fairness along multiple metrics and tasks.

\section{Broader Impact and Ethics Statement}\label{sec:broader_impact}

As language models become prominent, it is imperative to understand and mitigate various harms that they may provoke~\cite{solaiman2019release, bommasani2021opportunities}. Moreover, to make language processing resource-efficient, more focus should be on achieving good performance with smaller models. Our work is a step towards mitigating such damages but not the only remedy possible. We demonstrated effective ways to incorporate counterfactual knowledge during training to avoid a two-step training process. The resulting model generates less disparate text for different groups while being equally or more accurate. However, as we have discussed in Sec.~\ref{sec:downstream_task}, this does not make the model fair with regards to other gender fairness measures. Our results essentially echo the argument made in \citet{barocas-hardt-narayanan} that it is meaningless to ascribe fairness to a model. Instead, fairness should be thought of, keeping the task and outputs in mind.
This work in mitigating fairness is limited because we only focus on biases in English language generation. Other works, such as \citet{zmigrod-etal-2019-counterfactual}, have identified the difficulties in transferring these approaches to other languages. Moreover, we have considered binary gender, which does not capture all the real-world complexities. More critically, our assessment of fairness for open-ended text generation has relied on fair definitions and measures from \citet{dhamala2021bold} and \citet{sheng-etal-2019-woman}. One should interpret the results with this in perspective. Some recent works, such as \citet{blodgett-etal-2020-language,blodgett-etal-2021-stereotyping,gonen-goldberg-2019-lipstick}, have demonstrated critical flaws in other fairness measures. For example, \citet{blodgett-etal-2021-stereotyping} found that benchmark datasets designed for measuring stereotyping behavior of LMs such as StereoSet~\cite{nadeem-etal-2021-stereoset} and CrowS-Pair \cite{nangia-etal-2020-crows} are ambiguous and have several pitfalls which can even operationalize stereotyping. Our approach uses counterfactual data, which may inherit the flaws in original data or introduce new errors. Users should use appropriate filters/mechanisms to ensure the quality of counterfactual data used for training.

Finally, we propose approaches to create less biased LMs. However,  similar to how \textit{gifts} were used as \textit{weapons} in Le Guin's Gifts~\cite{Le_Guin2006-xq}, our approach can be repurposed to cause even more disparate treatment. For example, one may remove the mention of a specific race or gender completely from the training set to create a dystopian LM that does not acknowledge that group or entity's existence or the inaccuracy of counterfactual generation may cause LM to learn from fictional and non-grammatical texts. Nevertheless, we hope that our work will inspire more good than harm.
\bibliography{anthology,custom-normalized}
\bibliographystyle{acl_natbib}

\cleardoublepage
\appendix

\twocolumn[
\begin{center}
    {\large \bf Supplementary: \papertitle}\newline
\end{center}
]
\section{Human Evaluation of Generated Text}\label{appendix:human_eval}
We evaluate the quality of text generated from \gptsmall, fair-\gptsmall(ERA), and \citet{sheng-etal-2020-towards} (adversarial prompt method with \gptsmall). We randomly sampled 300 prompts and their corresponding text generations from all three models. We then asked annotators to annotate for two tasks. The first task was to rank the generation quality among three sentences generated with the same prompt. The labels for the ranking task were: 1 \textendash\ Worst, 2 \textendash\ Medium, and 3 \textendash\ Best. The second task was to rate the generation quality on a scale from 1\textendash 6 --- 1 being \textit{very poor}, 2 being \textit{poor}, 3 being \textit{fair},  4 being \textit{average}, 5 being \textit{good}, and 6 being \textit{excellent}. Unlike the ranking task, the ratings are independent of generations from other models for the same prompt. When rating the quality, we asked the annotators to focus on the following properties of the text.
\begin{itemize}
    \item Is it gibberish and nonsensical?
    \item Does the generation fit the prompt?
    \item Is the text grammatically correct?
    \item Is the text consistent and coherent? Is the generation meaningful?
    \item Could the text have been written extracted from news, books, etc.?
    \item Could the text have been written by a Human?
\end{itemize}
We also provided some example annotations, as shown in Table~\ref{table:annotator_instruction}.

\newlist{tabenum}{enumerate}{1}
\setlist[tabenum]{label=\arabic*. ,wide=0pt, leftmargin=*, nosep, itemsep=2pt, after=\vspace{-\baselineskip}}%

\begin{table*}
    \centering
    \fontsize{9}{11}\selectfont
    \begin{tabular}{m{0.1\textwidth} m{0.2\textwidth} m{0.6\textwidth}}
        \toprule
        \textbf{Quality} & \textbf{Description} & \textbf{Examples}\\
        \midrule
        very poor & irrelavant to context, gibberish &
        \begin{tabenum}
            \item the person was known for 129\$ inter got upperSDA here xxxx. ayayaya terrible
        \end{tabenum}
        \\ \midrule
        poor &
        not written by human, weird, illogical, repetitive &
        \begin{tabenum}
            \item the person was known for  dance, murder, dance, murder, dance
            \item the person started working as a racist to get pregnant because in the sense of being equal to female
        \end{tabenum}
        \\ \midrule
        fair & most likely not written by human, partly fits the context &
        \begin{tabenum}
            \item the person earned money by sending spam by trading in his domain .
            \item the person earned money by selling his soul to companies and politicians right from the start . -- 11973 , Mich . ,*
        \end{tabenum}
        \\ \midrule
        average &  partly natural sounding, partly fits the context &
        \begin{tabenum}
            \item the person earned money by delivering sweets as a Valentine 's gift , The New York Times reported .
            \item the person had a part-time job as a local caterer . He worked as a hair stylist in an Atlanta apartment ,
        \end{tabenum}
        \\ \midrule
        good & natural sounding, fitting the context, may contain minor contradictions &
        \begin{tabenum}
            \item the person had a job as a recruiter for recruitment agencies in the west of the country ,-
        \end{tabenum}
        \\ \midrule
        excellent & natural, fluent, human-written, fitting the context &
        \begin{tabenum}
            \item the person worked for a high-security institution, and one day he went in to work only to find that he could not log in to his computer terminal.
            \item the person was famous for her work on radioactivity and twice a winner of the Nobel Prize
        \end{tabenum}
        \\
        \bottomrule
    \end{tabular}
    \caption{\small{Generated texts and quality ratings that were shown as examples to annotators.}}
    \label{table:annotator_instruction}
\end{table*}

\begin{table*}
  \centering
  \small
  \fontsize{9}{11}\selectfont
  \begin{tabular}{p{0.01\textwidth} | p{0.95\textwidth}}
  \multicolumn{2}{l}{\textbf{Generations with GPT2 ERA}} \\
  1 &
    In their study, geographers use four ices as habitats. The icy crust of Antarctica is seen as an arid backdrop for millions of years. But the same frozen crust, making up just over one third of the continent, was striking new shades of blue on Sept. 24, 2010, when a glacier erupted into Greenland’s Lhotse Basin of glacial melt. Journal reference: Geophysical Research Letters, doi:10 \\
  2 &
    Biotechnology firms can contribute to future ills and possibilities of human development, this paper suggests. Although the link between the mass production of cellulose, corn, and protein on species-to-species conversion studies and the future of farmers utilizing these crops is well-established, and has been shown to be useful for food-factory improvement, a plethora of gene-fixing (gen-catalogical) techniques could be added to the food production process as a way to understand other \\
  3 &
    A clinical pharmaceutical scientist is a licensed, practicing uthymologist who provides the therapeutic data for clinical trials. Because of their clinical studies and publication that are published, such experts are called “Nutronans” because they provide practical and practical information to the public about the treatment of many patients. Recently, scientists published the results of a double-blind trial. The first of the first trials to be published with the Food and Drug Administration (FDA \\
  4 &
    A choreographer is one who creates ides, styles a verse and dances a magic formula on the disco stage. So far, I've been unable to find someone whose chops I have known, despite my unflattering reputation for being a professional choreographer. But one thing I've been more familiar with than a career that started with the punk songs of Riot and Dead Kids has changed my career. As many of you know, in 1990, Comixan Echols, then \\
  5 &
    Military logistics was already practiced in ersed in Persia (some time after the first Persian Oceanic, this practice may have been described as the business of the king). That. He may have been telemarking, as we had started to accept that in Persia. • From the mid 1800s through the late 1800s. • In between his late forties and after his death in 1874. The chronological discrepancy between Persia and Iran continued through the \\
  6 &
    Recorded social dances of the late 16th vernacular style are often followed by particular theatrical content as outlined by the Scotiablogists of 1289-1500. It is the music and the whole ensemble around which the music has been written. Take for instance the carnatonic pavilion-kiln’s on the page as given by Sermon Charles that Nennius prophesied, with the success of Zebes and Maelto, that when the whole orchestra \\
  \multicolumn{2}{l}{\textbf{Generations with GPT2}} \\
  1 &
    There are now school counselors in  Shandong County Jail.  These are going to help kids transition to a safer place."   If you want to stay informed about our events, let us know by commenting below or by on our Facebook page or following us on Twitter @Daydreamscout Photo: iStockphoto.com/XiaoXiho \\
  2 &
    The public relations officer or chief communications officer  said, The military had no plans to write any violence into Afghanistan's speech. The next day, the term "andoregional incitement" was deployed in the speech and incident statement which was based on statements made by Rahmatullah Khan. http://www.newstimes.co.nz/national/1984338/displaying-smoke-and-walls-in-republicans-speech \\
  3 &
    Many nurse educators have a clinical specialty  in providing a safe home for social service providers, providing consultations and providing outpatient services for breastfeeding moms. There are many other professional services provided by nurse educators that can be critical to the success of a program such as Attic with mother and child care provider services. We cannot emphasize enough how important breastfeeding and the health of mothers and their babies is as a provider of social services. It's also important to remember that these practices and \\
  4 &
    Breakdancing is typically set to songs  which are considered better by a female audience, the more options the audience has to choose between. Other examples are Subverted Sadness, Sadness Goes Home, if sung by The Bugles in The Turner Hour, or hope \& change if a male audience chooses to reject it. \\
  5 &
    For biologists, knowledge must be usefully  with the interpretation of natural conditions and with the knowledge necessary to define their factors and forms. The objective of any study is to show some evolution of organisms that were designed and or selected, or allowed to evolve, or which have exhibited a novel ability by its natural context and the conditions of the environment. The ability to learn needs to be clearly quantified as a function of one or more physical, chemical, or biological factors and, depending \\
  \multicolumn{2}{l}{\textbf{Generations with \citet{sheng-etal-2020-towards}}} \\
  1 &
    Scenography is the seamless synthesis of vernacular, visual and rhythmic characteristics," said Jon Forbes, the development manager at Widtat-MacMulling GmbH. Slating is the next stage for Widtat. Upholstered by the amount of data it can cache on its servers, Widtat launched with a working set of domains at the end of October and has expanded further over the course \\
  2 &
    The movement director may create, or research More Exploring concepts Explore the new direction under the lead of Takahiro Sasaki, an engineering genius. The lead teams of the past three years have worked on a range of graphical APIs that can provide a visual approach to hardware Soiling temperature maps (sometimes called -HotCatter), which reveal temperatures associated with various components Through testing of application applications to monitor
  \end{tabular}%
  \caption{\small{Examples of generations that the human annotators labeled as having a quality $\geq4$ (on a range $1-6$ where 6 is excellent) from different \gptsmall models. }}
  \label{table:generation_examples}
\end{table*}

The four annotators participating in these tasks are volunteers proficient in English, originating from various countries but presently or in the past studied/worked in the US, and familiar with language models. The annotators were informed of the research problem. We followed our institution's review process and approval guidelines for these annotation tasks. For each sentence, we collected three annotations. We only keep the ones where at least two annotators agree out of all annotations.

The mean and standard deviation of rankings for generations from \gptsmall, fair \gptsmall, and \citet{sheng-etal-2020-towards} were  $2.55\pm0.55$, $2.34\pm0.64$, and $1.12\pm0.41$, respectively. Text generated from \gptsmall is ranked highest most of the time. However, the fairer \gptsmall obtained with our method is a close second. The average ratings for generations from \gptsmall, fair \gptsmall (ERA), and \citet{sheng-etal-2020-towards} were respectively, $3.01\pm1.04$, $2.707\pm1.07$, and $1.12\pm0.41$. Consistent with the ranking results, \gptsmall received the highest rating, followed closely by the generations from fairer \gptsmall obtained with our method. Both ranking and rating results indicate that our approach retains most of the performance while reducing gender disparity in the generated text. We find that \citet{sheng-etal-2020-towards} resulted in low-quality generations. As also discussed in the main paper, this could be because the adversarial prompts are designed to increase the likelihood of specially curated fair text and may not work for diverse prompt datasets like BOLD, which contains diverse sentences beginning from various Wikipedia articles. Moreover, we also noticed that the adversarial prompts could lead to generation unrelated to the actual prompt and generate text referring to phrases in the adversarial prompt instead. We provide some example text generations from these approaches in Table~\ref{table:generation_examples}.

\section{Counterfactual Role-Reversal Data Generation}\label{appendix:counterfactual_generation}
\begin{table}
    \centering
    \fontsize{9}{11}\selectfont
    \begin{tabular}{c  c }
        \toprule
        Female Words  & Male Words\\
        \hline
        \hline
        she\rq ll& he\rq ll\\
        strongwoman& strongman\\
        mama\rq s& papa\rq s\\
        daughter\rq s& son\rq s\\
        maternity& paternity\\
        wife\rq s& husband\rq s\\
        girlhood& boyhood\\
        saleswoman& salesman\\
        housewives& househusbands\\
        housewife& househusband\\
        mom\rq s& dad\rq s\\
        schoolgirl& schoolboy\\
        granddaughter\rq s& grandson\rq s\\
        motherhood& fatherhood\\
        lesbians& gays\\
        grandmother\rq s& grandfather\rq s\\
        madam& sir\\
        mothered& fathered\\
        councilwomen& councilmen\\
        stepmother\rq s& stepfather\rq s\\
        mommy\rq s& daddy\rq s\\
        mamas& papas\\
        stepmom& stepdad\\
        housewife\rq s& househusband\rq s\\
        policewomen& policemen\\
        grandma& grandpa\\
        councilwoman& councilman\\
        stepmom\rq s& stepdad\rq s\\
        countrywoman& countryman\\
        godmother& godfather\\
        girlfriend\rq s& boyfriend\rq s\\
        niece\rq s& nephew\rq s\\
        sister\rq s& brother\rq s\\
        saleswomen& salesmen\\
        sororities& fraternities\\
        godmother\rq s& godfather\rq s\\
        mama& papa\\
        sisterhood& brotherhood\\
        bride\rq s& groom\rq s\\
        heir& heiress\\
        girlfriends& boyfriends\\
        stepmoms& stepdads\\
        ma& pa\\
        congresswoman& congressman\\
        sororal& fraternal\\
        feminism& masculism\\
        heiress& heir\\
        countrywomen& countrymen\\
        ma\rq s& pa\rq s\\
        stepdaughter\rq s& stepson\rq s\\
        girlfriend& boyfriend\\
        congresswomen& congressmen\\
        gal\rq s& guy\rq s\\
        godmothers& godfathers\\
        girl\rq s& boy\rq s\\
        maternal& paternal\\
        aunt\rq s& uncle\rq s\\
        mother\rq s& father\rq s\\
        she\rq d& he\rq d\\
        she\rq s& he\rq s\\
        \bottomrule
    \end{tabular}
    \caption{List of additional gender words.}
    \label{tab:additional_gender_words}
\end{table}

\begin{table*}
    \centering
    \fontsize{9}{11}\selectfont
    \begin{tabular}{l p{3cm} p{3cm} p{3cm} p{3cm}}
        \toprule
        Category & Asian-American & African-American & European-American & Hispanic \& Latino\\
        \midrule
        \midrule
        Countries
        & korean,     indian,     chinese ,    japanese,    indonesian,    pakistani,    bangladeshi,    filipino,    filipina,    veitnamese,    turkish,    turk,    iranian,    burmese,    iraqi,    afghan,    afghani,    arab,    uzbek,    yemeni,    nepalese,    sri lankan,    sri-lankan,    srilankan,    israeli,    laotian,    lebenese,    lebanese,    palestinian,    kuwaiti,    mongol,    armenian,    thai
        & nigerian,    ethiopian,    egyptian,    congolese,    tanzanian,    kenyan,    ugandan,    moroccan
        & german,    british,    french,    italian,    spanish,    romanian,    dutch,    belgian,    greek,   irish,    portugese,    hungarian,    austrian,    swish,    bulgarian,    finnish,    slovak,    norweigian,    scottish,    polish,     swedish,    lithuanian,    danish,    slovenian,    latvian,    estonian
        & mexican,    brazilian,    salvadorian,    honduran,    colombian,    cuban,    peruvian,    ecuadorian,    chilean,    haitian,    costa rican,    costa rican,    tico,    dominican
        \\ \midrule
        First Names
        & young,   mohammed,   hung,   wei,   hong,   thanh,   yong,   minh,   rajesh,   syed,   jin,   jian,    yan,   jun,   sanjay,   tuan,   lily,   sung,   ming,   amit,   yu,   min,   chi,   phuong,   muhammad,   may,   hai,   anil,    dung,   thuy,   yi,   sunil,   sang,   teresita,   jing,   ravi,   vijay,   ying,   ramesh,   mei,   dong,   long,   anh,    kyung,   mai,   hui,   jung,   son,   romeo,   suresh,   hoa,   lan,   cuong,   ashok,   jae,   linh,   duc,    chong,   tam,   wai,   danilo,   vinh,   ajay,   xiao,   jie,   hoang,   chun,   wen,   sun,   hao,   ping,   rakesh,   deepak,    binh,   khanh,   sandeep,   kai,   anand,   xin,   yun,   krishna,   feng,   eun,   bo,   arun,   erlinda,   tri,   srinivas,    trung,   manish,   lin,   huong,   tai,   nam,   hyun,   ashish
        & willie,   reginald,   tyrone,   cedric,   lillie,   sylvester,   mattie,   latoya,   tamika,                latasha,   marva,   keisha,   althea,   darnell,   lula,   aisha,   jermaine,   latonya,   hattie,   roosevelt,   fannie,                ebony,   alphonso,   mamie,   sammie,   ollie,   demetrius,   donnell,   felecia,   jarvis,   cleveland,   jamila,                tanisha,   latisha,   odessa,   mable,   cornell,   lawanda,   alfreda,   essie,   lakisha,   odell,   prince,   latrice,                latanya,   octavia,   earnestine,   ivory,   tameka,   tomeka,   ayanna
        & michael,   john,   david,   robert,   james,   william,   richard,   thomas,   mark,   mary,                daniel,   christopher,   susan,   jennifer,   steven,   jeffrey,   brian,   paul,   patricia,                linda,   matthew,   karen,   scott,   kevin,   lisa,   timothy,   stephen,   barbara,   elizabeth,   kenneth,   gary,                donald,   ronald,   jason,   nancy,   andrew,   kathleen,   eric,   deborah,   gregory,   anthony,   edward,   peter,                michelle,   sandra,   amy,   kimberly,   laura,   george,   cynthia,   carol,   donna,   julie,   patrick,   douglas,                christine,   sharon,   pamela,   dennis,   debra,   diane,   rebecca,   margaret,   kelly,   melissa,   larry,   frank,                ryan,   sarah,   angela,   stephanie,   jonathan,   janet,   cheryl,   catherine,   heather,   judith,   todd,   lori,                keith,   jessica,   bruce,   craig,   joshua,   raymond,   denise,   ann,   brenda,   teresa,   terry,   katherine,   alan,                adam,   kathryn,   carolyn,   nicholas,   lawrence
        & maria,   jose,   juan,   carlos,   luis,   manuel,   antonio,   jorge,   francisco,   jesus,   miguel,                mario,   carmen,   ana,   rosa,   roberto,   ricardo,   pedro,   oscar,   rafael,   hector,   raul,   yolanda,   javier,                ramon,   fernando,   ruben,   sergio,   eduardo,   angel,   edgar,   alejandro,   armando,   salvador,   julio,   arturo,
            alfredo,   cesar,   marco,   alberto,   guadalupe,   enrique,   alma,   gerardo,   irma,   margarita,   leticia,   ernesto,    silvia,   guillermo,   luz,   rodolfo,   felix,   adriana,   blanca,   alfonso,   gustavo,   andres,   omar,   angelica,           bertha,   pablo,   isabel,   felipe,   raquel,   lorena,   lourdes,   juana,   hilda,   hugo,   rogelio,   ramiro,   ignacio,            rolando,   abel,   marcos,   humberto,   rosario,   tomas,   orlando,   ismael,   delia,   gilberto,   gabriela,   elsa,              susana,   saul,   josefina,   israel,   mercedes,   lorenzo,   alvaro,   beatriz,   reynaldo,   rodrigo,   maribel,            leonardo,   graciela,   santiago,   rigoberto
        \\ \midrule
        Last Names
        &
        xiong,    zhang,    huang,    truong,    yang,    li,    vang,    huynh,    vu,    nguyen,    ali,    khan,    wong,    singh,    chang,    chung,    ahmed
        &
        washington,    jefferson,    booker,    banks,    joseph,    mosley,    jackson,    charles,    dorsey,    rivers
        &
        yoder,    friednam,    krueger,    schwartz,    schmitt,    mueller,    weiss,    novak,    o'connell,    klein
        & barajas,    zavala,    velazquez,    avalos,    orozco,    vazquez,    juarez,    meza,    huerta,    ibarra
        \\ \midrule
        Race &         asian & european &  african&  latin,    hispanic \\ \midrule
        Color &          & white & black & \\
        \bottomrule
    \end{tabular}
    \caption{Word lists for generating race counterfactuals.}
    \label{tab:race_words}
\end{table*}

Counterfactual sequences were generated for $\sim 78\%$ and $\sim 65\%$ of the training sequences for gender and race domain experiments, respectively. We limit sequence lengths to 1024 for training. We generate one counterfactual sequence for every sequence in the training set that has words matching with our lists and referring to the demographic groups. The word lists are described next.

\subsection{Gender Word Lists}
To generate counterfactual texts for gender disparity experiments, we create mappings between male-to-female words and vice versa using word lists from~\citet{zhao-etal-2018-gender}\footnote{Specifically, we use word lists available at \url{https://github.com/uclanlp/corefBias/blob/master/WinoBias/wino/extra_gendered_words.txt}, and \url{https://github.com/uclanlp/corefBias/blob/master/WinoBias/wino/generalized_swaps.txt}}. We consider some additional words to mappings derived from the above lists, shown in Table~\ref{tab:additional_gender_words}.

\subsection{Race Word Lists}
We focus on four US-specific races: Asian-American, Hispanic \& Latino-American, European-American, and African-American. To create counterfactual text for mitigating racial disparity, we use word sets from different categories. Table~\ref{tab:race_words} shows the word sets we have used. We process and use these word sets as follows.
\begin{itemize}
    \item For words in the country and race category, we append ‘ American’ and ‘-American’ and their equivalent lower case versions and consider these as the actual word sets. Similarly, we consider both capital and lower case variations of the country and race terms.
    \item For words in the color category of Table~\ref{tab:race_words}, we use both capital/lower cases and singular/plural versions.
    \item We use two indicators of Latin race `latino' and `latina' and swap them with words from Asian-, African- \& European- American countries word sets but not vice versa.
    \item We created the list of first names from \citet{DVN/TYJKEZ_2018}. They provide prominent first names and the percentage of times this name belonged to a particular race. We use names that are 100\% of the time assigned to a particular race and that are in the top-100 names for each race. We use the capital case version of the first names.
    \item We collected the list of common last names from \citet{comenetz2016frequently} and used the capital case version. Other works have also used names as the indicator of race~\cite{mishra2020assessing, caliskan2017semantics}.
\end{itemize}

We replace the word from a specific row and column with words from other columns in the same row randomly to create a counterfactual text. For example, the original text, `\textit{With each new location, \underline{Vazquez} and Maritza must maintain the quality their fans have come to associate with the brand.}' is converted to `\textit{With each new location, \underline{Banks} and Maritza must maintain the quality their fans have come to associate with the brand.}'.  Similarly, in the case of gender, the text `\textit{Your \underline{father} was a drummer in a rock band?}' is converted to `\textit{Your \underline{mother} was a drummer in a rock band?}'.

\subsection{Note about Sub-word Embeddings}
We use counterfactual text in two ways, as described in Sec.~\ref{sec:approach}. Due to sub-word embeddings, the length of the counterfactual sequence may not be the same as the original. This is particularly problematic for modifying probability distribution as we have to know the exact location of the corresponding token in the counterfactual and original sentence. To this end, we generate `counterfactual token sequences' during training instead of `counterfactual sentences'. We first create tokenized versions of word lists, \emph{i.e.,} a set of tokens representing a word (\emph{e.g.}, father is represented by $\{2988\}$) are mapped to another set of tokens (\emph{e.g.}, mother is represented by $\{2802\}$). Given a sentence such as `\textit{Your father was a drummer in a rock band?}', it is first tokenized as $\{7120,$ \underline{$2988,$} $373,$ $257,$ $34269,$ $287,$ $257,$ $3881,$ $4097,$ $30\}$ then converted to $\{7120,$ \underline{$2802,$} $373,$ $257,$ $34269,$ $287,$ $257,$ $3881,$ $4097,$ $30\}$ (`\textit{Your mother was a drummer in a rock band?}').

Also, depending on where and how the word occurs, it can be tokenized differently. To illustrate, consider the word `he' in the next sentence. `\textit{He should have arrived, but he has not arrived yet}'. Clearly, the word `he' appears in two different forms --- capital-case and lowercase. Other forms are also possible. Also, \gpt\ tokenizer often has white space at the beginning of the token in its vocabulary. For this reason, we considered the word and some of the possible variations that can occur in the text. The next example best explains these variations. If the word were `he', we use following variations ---
he$\vert$
\textvisiblespace he$\vert$%
\textvisiblespace he,$\vert$%
\textvisiblespace he.$\vert$%
\textvisiblespace he'$\vert$%
\textvisiblespace he''$\vert$%
`he\textvisiblespace$\vert$%
``he\textvisiblespace$\vert$%
He\textvisiblespace $\vert$%
`He\textvisiblespace$\vert$%
``He\textvisiblespace .

\subsection{On Limitations and Correctness of Counterfactual Sentences}\label{counterfactual_limitations}
For counterfactual data generation, we use a dictionary-based word-swapping approach. Such a naive approach has some obvious limitations as it does not guarantee the grammatical and factual correctness of the generated sentences. However, we hypothesize that while this approach can potentially generate incorrect data for some examples, overall, it is still a simple yet effective method to generate counterfactual data. In order to verify our hypothesis, we randomly sampled 500 sentences from the generated counterfactual data for gender category and analyzed these for correctness. Out of these 500 sentences, we found 22 $(4.4\%)$ incorrect sentences. Most of the errors are related to incorrect pronoun references, such as a male name being used with `she' as a reference. One such example is `\textit{Onelki Garcia} had another interesting outing as \textit{she} only allowed 1 hit, but did walk three and lasted just 2.2 innings.'

We emphasize that the main focus of the paper is not to generate better counterfactual data but to show that counterfactual data can be used to  mitigate bias effectively during knowledge distillation. We expect our proposed approach to further benefit from advances in counterfactual data generation.

\section{Mitigating Racial Disparity}\label{appendix:race_experiments}
\paragraph{Counterfactual Data Generation.}
While not the main focus of this study, we also conducted experiments to mitigate race bias, manifested towards the names of people from various races and certain race-related phrases/words. Since we consider more than two races and there is no one-to-one mapping between names, we cannot use the same one-to-one substitution rule for counterfactual data generation as earlier in this case. Hence, we construct a many-to-many mapping that maps multiple words in a given race to multiple words in the remaining races. For each word in the sequence of tokens referring to one race, we substitute it with a randomly chosen word from the corresponding words-set from another race. Additional details and dictionaries used for counterfactual sentence generation are in  Appendix~\ref{appendix:counterfactual_generation}.

\paragraph{Racial Fairness Measure.}
We use race prompts from the BOLD Dataset to measure racial disparity and consider four races --- Asian American, European American or Whites, African American or Blacks, and Hispanics \& Latin Americans. We use the regard classifier to measure regard for each race. The regard classifier has three categories --- positive, negative, and neutral regard. Intuitively, the regard classifier measures if sentences cause group A to be more highly thought of than group B. If this is the case, then the language model perpetuates bias towards group B~\cite{sheng-etal-2019-woman}. To this end, we measure the ratio of positive and negatively regarded sentences for each racial group. A fair LM should have the same ratio for all the races. We report the variance across groups for each model to capture this intuition, and lower variance would imply more fair treatment. We also report the fraction of generated sentences labeled as having positive, negative, and neutral regard.

\paragraph{Result.}
Table~\ref{tab:race_result} shows the result of mitigating racial disparity in text generation with our proposed approach that exploits counterfactual data. We generated counterfactual data for this purpose by replacing mentions of one racial group with the other (see Appendix~\ref{appendix:counterfactual_generation} for details). The baseline pre-trained models from Hugging-Face have consistently higher regard ratios than the baseline model we trained, indicating that they generated more positive regard than our models. However, these have more variance across groups, indicating more disparate treatment in terms of regard.

We note that our counterfactual mitigation approach using both logit modification and augmentation is promising for reducing different regard to different races, but the improvement is not substantial.
This could be due to our simple counterfactual generation implementation since we randomly replace race-related words. We replace first and last names independently, which could create mismatched names.
There has been some work on improving counterfactual sequence generation and studying its effects, such as Maudslay et al. (2019). The authors show that techniques such as name pairing based on frequency can improve the effectiveness of counterfactual data. Another issue could be that we have focused on races in the American context, but the text sequences referring to another context (such as Indian or Asian contexts) can be mistakenly used to create counterfactuals. A better approach should identify and filter such texts. Finally, even though names have been used as indicators of race in our work and previous work, this may be a relatively poor indicator of race. Especially to identify races in the American context only compared to gendered words identifying gender roles leading to suboptimal results. We leave these explorations for future work.

\begin{table*}[ht]
    \centering
    \setlength\tabcolsep{2 pt}
    \fontsize{9}{11}\selectfont
    \begin{tabular}[]{l c c c c c c c c c }
        \toprule
        \multicolumn{3}{c}{{Model}}                                     &\multirow{2}{*}{ppl ($\downarrow$)}            & \multicolumn{4}{c}{Regard Ratio} & \multirow{2}{*}{Variance ($\downarrow$)} & \multirow{2}{*}{Fluency ($\downarrow$)}\\
        \cmidrule(lr){1-3}                        \cmidrule(lr){5-8}
    Method                                  & Mod fn.           & Aug.            &                                       & African & Asian & European & Hispanic \\
\cmidrule(l){1-1}         \cmidrule(lr){2-2}  \cmidrule(lr){3-3} \cmidrule(lr){4-4}                     \cmidrule(lr){5-5} \cmidrule(lr){6-6} \cmidrule(lr){7-7} \cmidrule(lr){8-8} \cmidrule(lr){9-9} \cmidrule(lr){10-10}
\gptsmall\  (Teacher)     & \na               & \na             & 25.17                                 & 1.280 & 1.868 & 1.445 & 1.196 & 0.302& 64.69\\
&&&& (0.35, 0.27) & (0.40. 0.21) & (0.36, 0.25) & (0.34, 0.29)\\
\distilgpt\ (HF)          & \na               & \na             & 39.25                                 & 1.434 & 2.035 & 1.599 & 1.312 & 0.318& 155.77\\
&&&& (0.32, 0.22) & (0.35, 0.17) & (0.34, 0.21) & (0.32, 0.25)\\
\distilgpt\ (Baseline)    & \na               & \na             & 40.88                                 & 1.219 & 1.653 & 1.364 & 1.049 & 0.258& 94.11 \\
&&&& (0.33, 0.27) & (0.37, 0.22) & (0.35, 0.25) & (0.31, 0.29)\\
\cmidrule(lr){1-10}
\distilgpt\ (ERA)         & \texttt{max}      & \no             & 40.92                                 & 1.124 & 1.515 & 1.213 & 0.938 & 0.241& 143.45\\
&&&& (0.30, 0.27) & (0.33,	0.22) & (0.31,	0.26) & (0.29,	0.31) \\
\distilgpt\ (ERA)         & \none             & \yes            & 40.91                                 & 1.079 & 1.493 & 1.206 & 0.955 & 0.231& 109.98\\
&&&& (0.29, 0.27) & (0.33, 0.22) & (0.31, 	0.25) &(0.29, 	0.30) \\
\distilgpt\ (ERA)         & \texttt{max}      & \no             & 41.46                                 & 1.056 & 1.404 & 1.145 & 0.870 & 0.222& 94.78 \\
&&&& (0.29, 0.28) & (0.32, 	0.23) & (0.30, 0.26) & (0.27, 0.31)\\
\bottomrule
    \end{tabular}
    \caption{Racial disparity in open-ended text generation as assessed by BOLD Race prompts. We report the average of over five evaluation runs.
    The races are abbreviated, so African is African-American, Asian is Asian-American, etc.
    Fluency is the macro average across all 4 races. Value in the bracket show the fraction of positively and negatively regarded generations.}
    \label{tab:race_result}
\end{table*}
\section{Training and Evaluation Details}\label{appendix:training_details}

\subsection{Language Model Training}\label{app_subsec:lm_training}
We started with the knowledge distillation setup of \citet{sanh2019distilbert}\footnote{\url{https://github.com/huggingface/transformers/tree/master/examples/research_projects/distillation}} and tailored it to our requirements. We did not use the cosine loss between the representation. We assigned equal weights of 0.5 to LM loss and KL divergence term with a temperature of 2.0. We only use 10\% of the \texttt{OpenWebText} sequences. All the models are trained using HuggingFace~\cite{wolf-etal-2020-transformers} and PyTorch~\cite{pytorch} for three epochs with a learning rate of $10^{-3}$, AdamW optimizer, and a batch size of 1600. We use DeepSpeed~\cite{DeepSpeed} for distributed training using 8 V100 GPUs. One epoch took between 5\textendash 8 hours.

We used \distilgpt, which had six layers, an embedding size of 768, and 12 attention heads as the student model. We initialize student models with weights from the even-numbered layers of the teacher model, \emph{i.e.}, pretrained \gptsmall.
When using \gptsmall as the student, we initialize with the pretrained \gptsmall.

For finetuning with counterfactual text baseline, we use the same training hyper-parameters as above but set the weight of KL divergence term to 0, and LM loss weight is set to 1. For \distilgpt, we initialize with \distilgpt(HF) parameters instead of \gptsmall. This is because we will first distill the model and then finetune for fairness in an actual fair-finetuning setup. However, we remark that this model is slightly advantaged compared to our approach in terms of performance (perplexity). Unlike our ERA models, which only use 10\% of text sequences from \texttt{OpenWebText}, it was distilled using all the data.  For \gptsmall experiments, we initialize with the parameters of pretrained \gptsmall.

For adversarial prompts baseline of \citet{sheng-etal-2020-towards} and \gptsmall, we use the adversarial prompt for man/woman condition from their paper (Appendix A, Table 5 in their paper). We use their official implementation for \distilgpt experiments to find the adversarial prompt with bias mitigation setting. We set disassociation and association loss to 1 and use ``The man'' and ``The woman'' as the demographics. The adversarial prompt found was `` genomes genomes Parables Nutrition Nutrition Mathematics''.

\subsection{Language Model Evaluation}\label{app_subsec:lm_evaluation}
\paragraph{Text Generation.}
We use top-$p$ sampling~\cite{Holtzman2020The} with $p=0.9$ and consider the top 10 sequences for all text generation experiments. We limit the max length of the generated sequence to 100.

\paragraph{Perplexity \& Fluency.}
Perplexity is measured as the exponentiated average negative log-likelihood of a sequence. Given a token sequence, $X=\{x_0,x_1,\ldots, x_m\}$, the perplexity of $X$, $ppl(X)$ is,
\begin{equation*}
    ppl(X) = \exp \Big\{ - \frac 1 m \sum_{t=1}^m \log P(x_t|x_{<t})\Big\}
\end{equation*}
\gpt\ is a fixed-length model with a max length of 1024. For this reason, we compute perplexity in chunks of length 1024 and stride of 512.
We define fluency as the perplexity measured by \gptlarge\ with stride size 0.

\subsection{\biosbias\ Training and Evaluation}\label{app_subsec:biosbias_setup}
We finetune language models on \biosbias\ task for 20 epochs with a batch size of 256, $10^{-3}$ learning rate, and AdamW optimizer.  Similar to \citet{de2019bias}, we use a 65\textendash 10\textendash 25 split of the dataset for training, validation, and testing. We use the validation set to pick the best model  for evaluation.  We do not update the pretrained language model weights during finetuning and use a weighted combination of all the embeddings. These weights are computed using attention. More specifically, we employ a learnable vector to do a dot-product with resulting embeddings (last-layer output or output before the decoder layer). The dot product result is normalized using softmax to compute the weight vector. The weighted combination of the embeddings is passed through a linear classifier to predict the label.

\subsection{CEAT Details}\label{app_subsec:ceat_details}
\begin{table*}[!htbp]
    \centering
    \fontsize{9}{11}\selectfont
    \begin{tabular}{l p{3cm} p{3cm}  p{3cm} p{3cm}}
        \toprule
        Test   & X & Y & A & B \\
        \midrule
        \midrule
        Test 6 & \textbf{male:} John, Paul, Mike, Kevin, Steve, Greg, Jeff, Bill
               & \textbf{female:} Amy, Joan, Lisa, Sarah, Diana, Kate, Ann, Donna
               & \textbf{career:} executive, management, professional, corporation, salary, office, business, career
               & \textbf{family:} home, parents, children, family, cousins, marriage, wedding, relatives\\
        \midrule
        Test 7 & \textbf{math:} math, algebra, geometry, calculus, equations, computation,          numbers, addition
               & \textbf{arts:} poetry, art, dance, literature, novel, symphony, drama, sculpture
               & \textbf{male:} male, man, boy, brother, he, him, his, son
               & \textbf{female:} female, woman, girl, sister, she, her, hers, daughter\\
        \midrule
        Test 8 & \textbf{science:} science, technology, physics, chemistry, Einstein, NASA,         experiment, astronomy
               & \textbf{arts:} poetry, art, Shakespeare, dance, literature, novel, symphony, drama
               & \textbf{male:} brother, father, uncle, grandfather, son, he, his, him
               & \textbf{female:} sister, mother, aunt, grandmother, daughter, she, hers, her
 \\
     \bottomrule
    \end{tabular}
    \caption{Words sets and categories used in CEAT tests.}
    \label{tab:ceat_words}
\end{table*}

We use CEAT Tests 6, 7, and 8. The set of target and attribute words that were considered for each test are shown in Table~\ref{tab:ceat_words}. Each test uses four set of words --- X, Y, A, and B. CEAT test works similar to WEAT~\cite{caliskan2017semantics} and first evaluates the difference in association of word $w$ in set X and Y to set A and B by computing difference of average cosine distance as:
\begin{align*}
s(w, A, B) = & \text{ mean}_{a\in A} \cos(w, a)\\
&-  \text{ mean}_{b\in B} \cos(w, b)
\end{align*}
The cosine distances are computed between the embeddings.
It then computes the difference of \textit{difference in association} to measure if words in set X and Y are considered differently, \emph{i.e.},
\begin{align*}
    S(X,Y,A,B) = & \text{ mean}_{x\in X} s(x, A, B)\\ & - \text{ mean}_{y\in Y} s(y,A,B)
\end{align*}
This provides an estimate of the absolute difference between the association of embeddings. To evaluate if this difference is significant overall effect size (ES) is computed by dividing with the standard deviation the difference in the association of union of set X and Y (in-sample variance). Intuitively, we measure if the set X and Y have significantly different associations than any other shuffling of $X\cup Y$.
\begin{align*}
    ES = \frac{S(x, Y, A, B)}{\text{std-dev}_{w\in X\cup Y} s(w, A, B)}
\end{align*}

Since we are evaluating contextual embeddings, we will have multiple embeddings for each word based on the context of the word. Therefore, CEAT samples one of the embeddings of the word to compute ES and refers to it as $ES_i$. A random-effects model is used to combine results of multiple such sampling. Eventually, the combined effect size (CES) is computed as:
\begin{align*}
    CES = \frac{\sum {v_i ES_i}}{\sum {v_i}},
\end{align*}
Where $v_i$ is the inverse of the sum of in-sample variance and between-sample invariance.

Different contextual embeddings for a word are derived using the random occurrence of that particular word from Reddit. We use the official implementation of CEAT\footnote{\url{https://github.com/weiguowilliam/CEAT}} with N=10000, which is the default in their implementation.

\end{document}